\documentclass[10pt, journal]{IEEEtran}
\newcommand{\subparagraph}{}
\usepackage{cite}
\usepackage{amsmath,amssymb,amsfonts}
\usepackage{graphicx}
\usepackage[font=footnotesize]{caption}
\usepackage{comment}
\usepackage{subcaption}
\usepackage{physics,amsmath}
\usepackage[ruled,vlined]{algorithm2e}
\include{pythonlisting}
\usepackage{float}
\usepackage{graphics}
\usepackage{titlesec}
\usepackage{subfig}
\usepackage{graphicx}
\usepackage{tabularx}

\usepackage{xcolor}
\usepackage{titlesec}
\usepackage{subfig}
\usepackage{graphicx}
\usepackage{mathtools}
\usepackage{cite}
\usepackage[font=footnotesize]{caption}
\usepackage{array}
\usepackage[english]{babel}
\usepackage[utf8x]{inputenc}
\usepackage{xcolor}
\usepackage[colorinlistoftodos]{todonotes}

\usepackage{enumitem}

\usepackage{amsmath}
\usepackage{amssymb}
\usepackage{amsfonts}
\usepackage{amstext}
\usepackage{amsthm}
\usepackage{soul}

\usepackage{algpseudocode}
\usepackage{tikz}
\newcommand{\ballnumber}[1]{\tikz[baseline=(myanchor.base)] \node[circle,fill=.,inner sep=1pt] (myanchor) {\color{-.}\bfseries\footnotesize #1};}
\titlespacing{\section}{0pt}{2ex}{1ex}
\titlespacing{\subsection}{0pt}{0.5ex}{0.2ex}
\titlespacing{\subsubsection}{0pt}{0.5ex}{0ex}
\begin{document}
\title{Containing Analog Data Deluge at Edge through Frequency-Domain Compression in Collaborative Compute-in-Memory Networks}
\author{Nastaran Darabi, and Amit R. Trivedi}
\author{Nastaran Darabi, and Amit Ranjan Trivedi \\ 
AEON Lab, University of Illinois at Chicago (UIC), Chicago, IL, USA}

\maketitle
\begin{abstract}
Edge computing is a promising solution for handling high-dimensional, multispectral analog data from sensors and IoT devices for applications such as autonomous drones. However, edge devices' limited storage and computing resources make it challenging to perform complex predictive modeling at the edge. Compute-in-memory (CiM) has emerged as a principal paradigm to minimize energy for deep learning-based inference at the edge. Nevertheless, integrating storage and processing complicates memory cells and/or memory peripherals, essentially trading off area efficiency for energy efficiency. This paper proposes a novel solution to improve area efficiency in deep learning inference tasks. The proposed method employs two key strategies. Firstly, a Frequency domain learning approach uses binarized Walsh-Hadamard Transforms, reducing the necessary parameters for DNN (by 87\% in MobileNetV2) and enabling compute-in-SRAM, which better utilizes parallelism during inference. Secondly, a memory-immersed collaborative digitization method is described among CiM arrays to reduce the area overheads of conventional ADCs. This facilitates more CiM arrays in limited footprint designs, leading to better parallelism and reduced external memory accesses. Different networking configurations are explored, where Flash, SA, and their hybrid digitization steps can be implemented using the memory-immersed scheme. The results are demonstrated using a 65 nm CMOS test chip, exhibiting significant area and energy savings compared to a 40 nm-node 5-bit SAR ADC and 5-bit Flash ADC. By processing analog data more efficiently, it is possible to selectively retain valuable data from sensors and alleviate the challenges posed by the analog data deluge.  
\end{abstract}
  
\begin{figure}[t!]
    \centering
    \includegraphics[width=\linewidth]{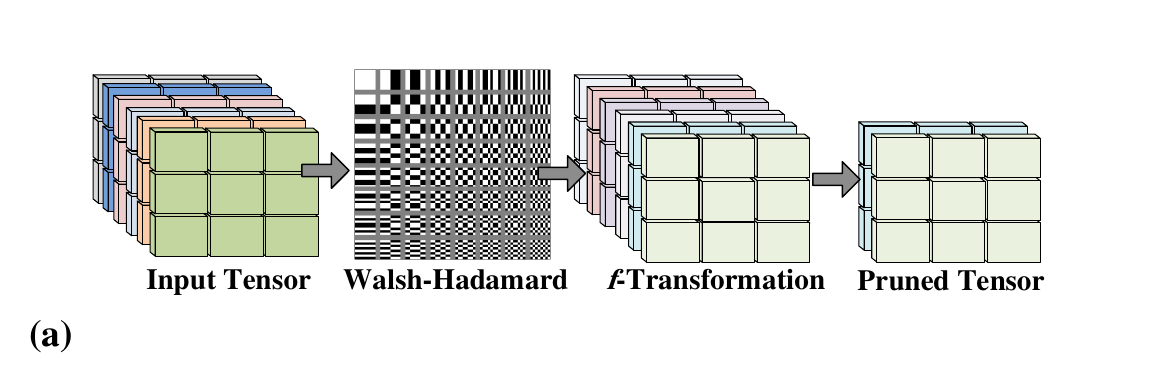}
    \includegraphics[width=0.98\linewidth]{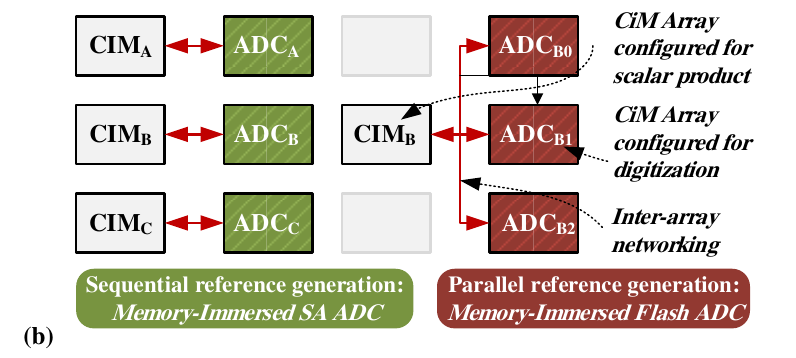}
    \includegraphics[width=0.49\linewidth]{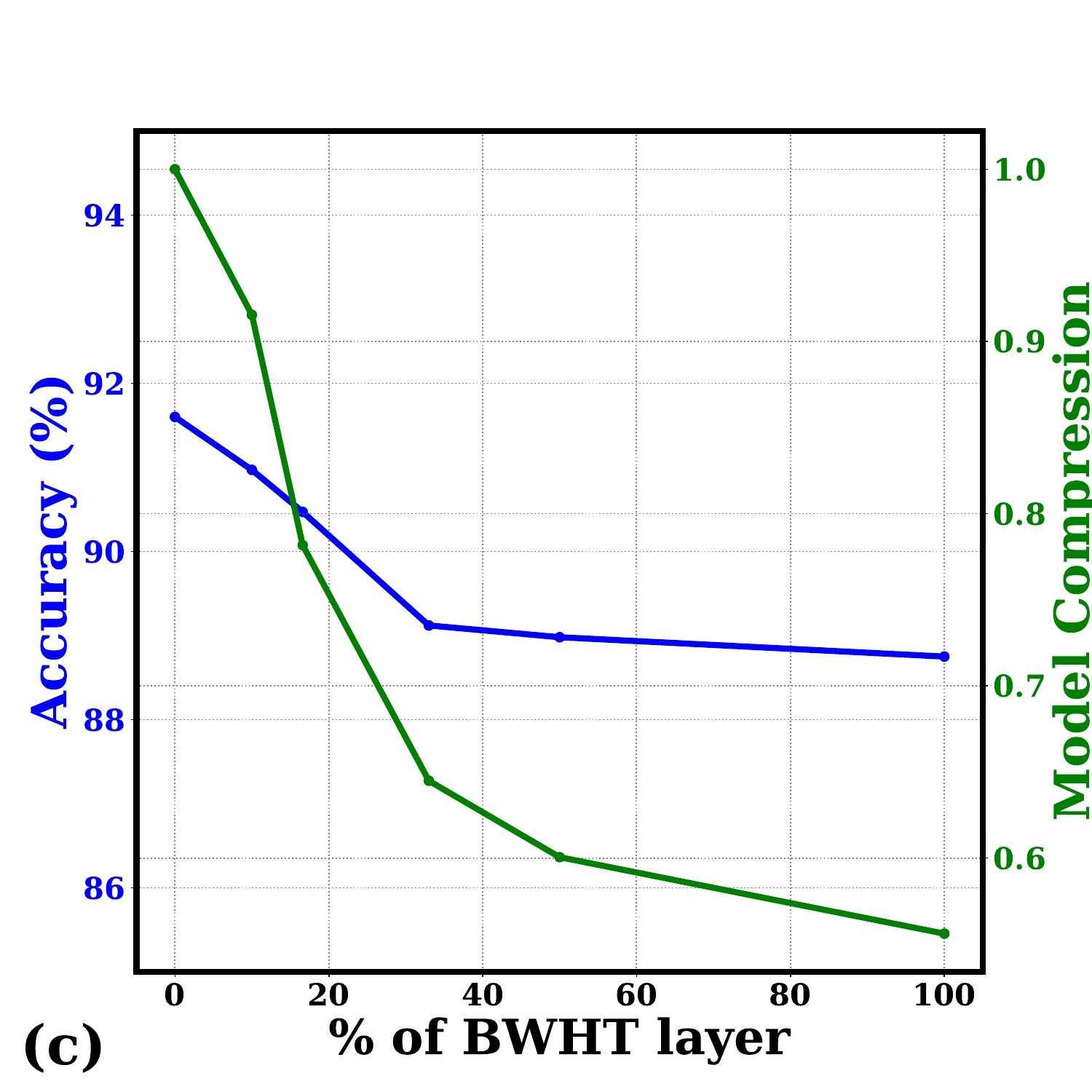}
    \includegraphics[width=0.49\linewidth]{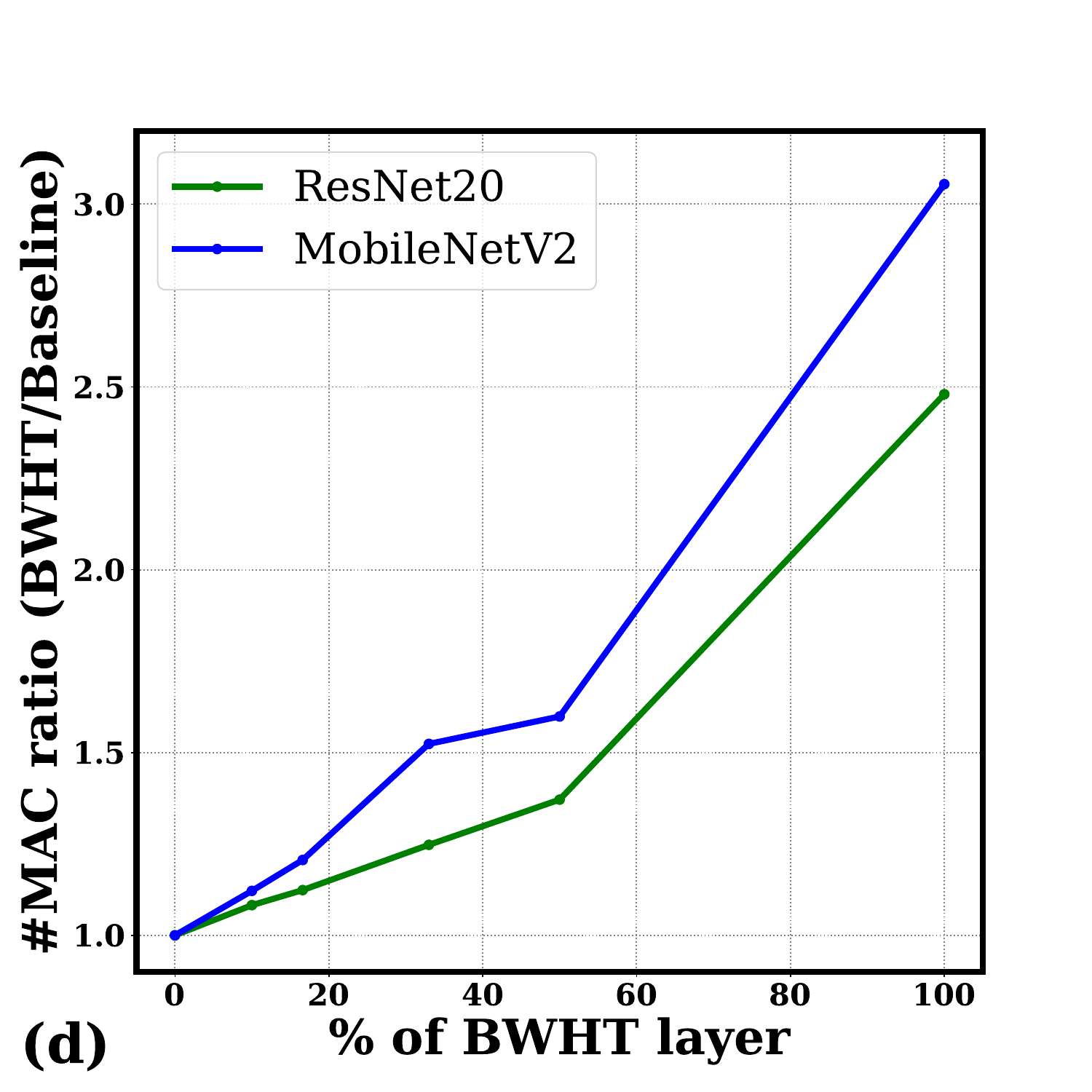}
    \caption{\textbf{Addressing the Data Deluge with Frequency-Domain Processing of Neural Networks and Memory-Immersed Collaborative Digitization:} The figure illustrates two key strategies for managing the deluge of data in neural networks and analog systems. \textbf{(a)}  Frequency-Domain Processing of Neural Networks: Frequency transformations of neural tensors are used to manage the data deluge.  \textbf{(b)}  Memory-Immersed Collaborative Digitization for Analog Data Deluge: The figure also illustrates how Compute-in-Memory (CiM) arrays are coupled for sequential (left) and parallel (right) reference generation for memory-immersed Analog-to-Digital Converters (ADC). These approaches are designed to manage the deluge of analog data by converting it into a digital format, making it more manageable and suitable for further processing. \textbf{(c)} The impact of processing increasing layers of ResNet20 with Walsh-Hadamard transforms (WHT) on prediction accuracy and model compression is shown. \textbf{(d)} The increase in multiply-accumulate (MAC) operations under frequency domain processing compared to conventional processing for MobileNetV2 and ResNet20 is also demonstrated.}
    \label{fig:Introduction}
\end{figure}

\begin{figure*}[t!]
    \centering
    \includegraphics[width=\linewidth]{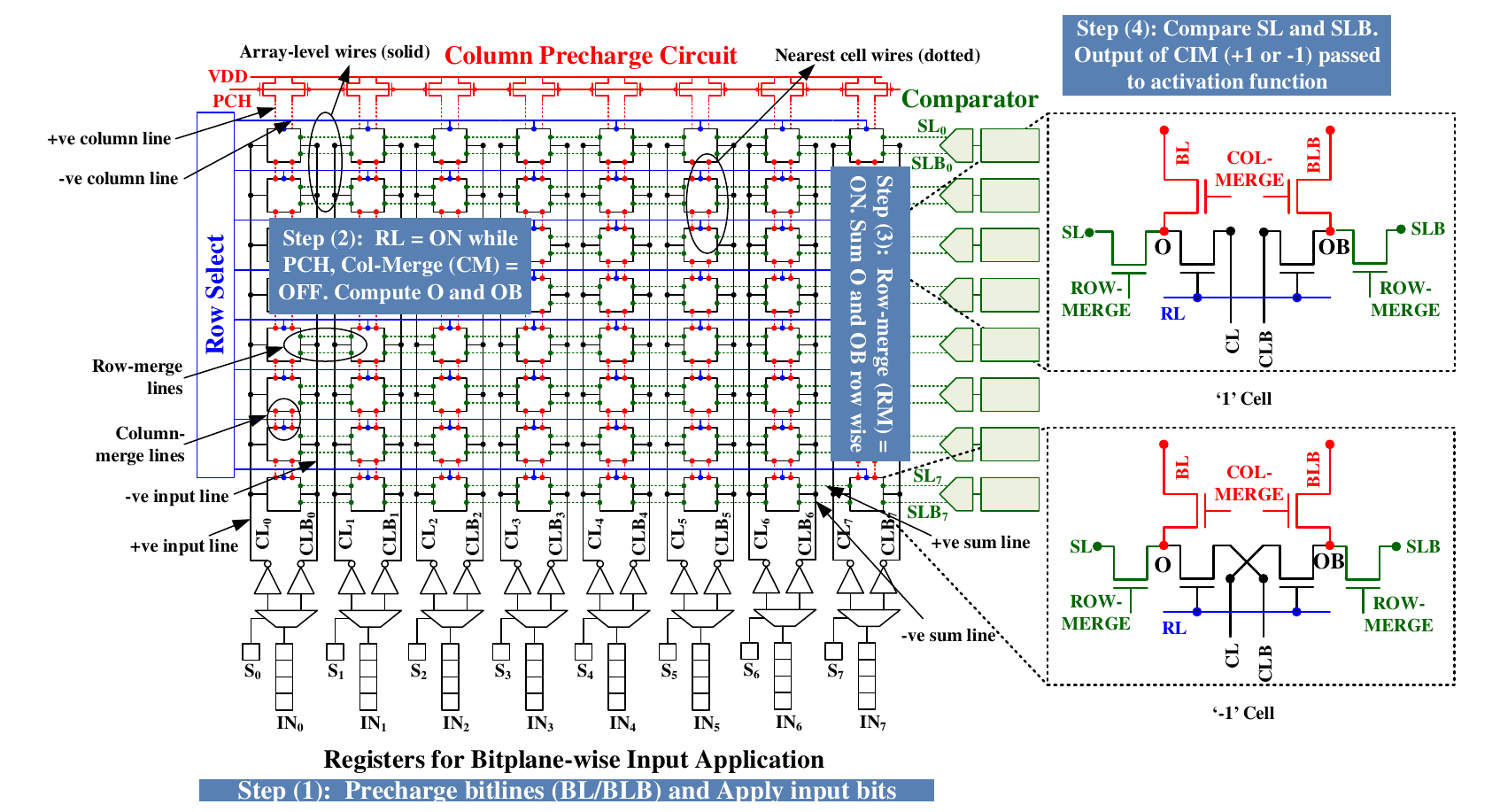}  
    \caption{\textbf{Architecture and operation flow for an analog acceleration of frequency-domain neural processing:} The operation consists of four steps: (1) precharging bit lines (BL/BLB) and applying input, (2) enabling parallel local computations in O and OB, (3) activating row-merge to connect all cells row-wise and summing O/OB in sum lines (SL/SLB), and (4) comparing SL/SLB values and applying soft thresholding for accurate output generation.}
    \label{fig:top-level architecture}
\end{figure*}

\section{Introduction}

The advent of deep learning and its application in critical domains such as healthcare, finance, security, and autonomous vehicles has led to a data deluge, necessitating efficient computational strategies \cite{chen2019deep,tayebati2023hybrid, hassanalieragh2015health}. Deep neural networks (DNNs), which are increasingly deployed at the network's edge, are particularly challenging due to their complexity and the limited computing and storage resources at the edge \cite{yu2020easiedge}. This work presents novel techniques to address these challenges, focusing on the compute-in-memory (CiM) processing of DNNs and frequency-domain model compression.

CiM integrates model storage and computations, reducing significant data movements between intermediate memory hierarchy and processing modules that hinder the performance of conventional digital architectures for DNNs. Traditional memory structures such as SRAM \cite{sehgal2021trends}, RRAM \cite{yu2021compute}, and embedded-DRAM \cite{jung2022dualpim,xie202116} can be adapted for CiM, making it an attractive scheme for cost-effective adoption in various systems-on-chip (SOC). CiM schemes leverage analog representations of operands to simplify their summation over a wire by exploiting Kirchhoff's law, thereby minimizing the necessary workload and processing elements \cite{hung20228, sakr2021signal, 9833492, 9972346, nasrin2021mf}.

However, the analog computations in CiM present significant implementation challenges, particularly the need for digital-to-analog converters (DAC) and analog-to-digital converters (ADC) to operate on digital inputs and digitize the analog output for routing and storage. This work proposes a novel memory-immersed digitization that can preclude a dedicated ADC and its associated area overhead, Fig. \ref{fig:Introduction}(b) The proposed scheme uses parasitic bit-lines of memory arrays to form within-memory capacitive DAC, and neighboring memory arrays collaborate for resource-efficient digitization \cite{Shamma_AICAS}. Although \cite{nasrin2021mf} similarly explored memory-immersed digitization, the presented results were based on simulations only, and only SA functionality was shown. Meanwhile, in this work, the proposed techniques are characterized on a test chip designed in 65 nm CMOS technology. We demonstrate 5-bit memory-immersed ADC operation using a network of 16$\times$32 compute-in-SRAM arrays. Compared to a 40 nm-node 5-bit SAR ADC, our 65 nm design requires $\sim$25$\times$ less area and $\sim$1.4$\times$ less energy by leveraging in-memory computing structures. Compared to a 40 nm-node 5-bit Flash ADC, our design requires $\sim$51$\times$ less area and $\sim$13$\times$ less energy. 

On the other hand, frequency-domain model compression, an efficient alternative to traditional model pruning techniques, leverages fast algorithms for transforms such as the discrete cosine transform (DCT) or discrete Fourier transform (DFT) to identify and remove redundant or uncorrelated information in the frequency domain \cite{chen2016compressing, liu2018frequency, xu2020learning, rossi2020walsh, pan2021fast, pan2022dct, koizumi2018end, 8203813, xu2019frequency, mohsen2018classification, xue2013restructuring, riera2022dnn}. This work focuses on Walsh-Hadamard transform (WHT)-based model compression, which can reduce model size with limited accuracy loss, shown in Fig. \ref{fig:Introduction}(c). However, as demonstrated in Fig. \ref{fig:Introduction}(d) it also introduces a notable increase in the required multiply-accumulate (MAC) operations, offsetting the benefits of model size reduction. To address this challenge, this work presents a novel analog acceleration approach\cite{darabi2023adc}.

These techniques and architectures present a promising solution for sustainable edge computing, effectively addressing the challenges posed by the analog data deluge \cite{sehgal2021trends, yu2021compute, jung2022dualpim, xie202116, 10129331, 9971009, hung20228, sakr2021signal, nasrin2021mf}. By improving area efficiency in deep learning inference tasks, reducing energy consumption, and leveraging output sparsity for efficient computation, these techniques enable better handling of high-dimensional, multispectral analog data \cite{chen2016compressing, liu2018frequency, xu2020learning, rossi2020walsh, pan2021fast, pan2022dct, koizumi2018end, 8203813, xu2019frequency, mohsen2018classification, xue2013restructuring, riera2022dnn}. This makes them a promising solution for sustainable data processing at the edge, paving the way for the next generation of deep learning applications in scenarios where area and power resources are limited \cite{chen2019deep, hassanalieragh2015health, yu2020easiedge}.

\begin{itemize}
    \item Firstly, we introduce a Compute-In-Memory (CIM) architecture that capitalizes on Blockwise Walsh-Hadamard Transform (BWHT) and soft-thresholding techniques to compress deep neural networks effectively. This innovative CIM design enables computation in just two clock cycles at a speed of 4 GHz, a feat made possible by implementing full parallelism. Notably, our design eliminates the need for ADCs or DACs. Furthermore, we propose an early termination strategy that exploits output sparsity, thereby reducing both computation time and energy consumption.
    \item Secondly, in scenarios where the inclusion of ADCs in our design becomes necessary, we present a memory-immersed collaborative digitization technique. This method significantly reduces the area overheads typically associated with conventional ADCs, thus facilitating the efficient digitization of large data volumes. This technique not only optimizes the use of space but also enhances the system's overall efficiency by streamlining the conversion process. It ensures that even with the inclusion of ADCs, the system maintains high performance while handling substantial data volumes.
\end{itemize}

Section II provides the essential background information required to understand the techniques proposed in this paper. Section III delves into the proposed design for a CMOS-based in-memory frequency-domain transform. In Section IV, we elaborate on the proposed memory-immersed collaborative digitization technique. Section V is dedicated to discussing the sustainability aspects of our design. Finally, Section VI concludes the paper, summarizing the key points and findings.

\section{Background and Related Works}
\subsection{Walsh-Hadamard Transform (WHT)}
The Walsh-Hadamard Transform (WHT) is akin to the Fast Fourier Transform (FFT) as both can transmute convolutions in the time or spatial domain into multiplications in the frequency domain. A distinguishing feature of WHT is that its transform matrix consists solely of binary values (-1 and 1), eliminating the need for multiplications, thereby enhancing efficiency.

Given $X, Y \in \mathbf{R}^m$ as the vector in the time-domain and WHT-domain, respectively, where \textit{m} is an integer power of 2 ($2^k , k\in \mathbf{N}$), the WHT can be expressed as:
\begin{equation}
Y = W_k X
\end{equation}
Here, $W_k$ is a $2^k \times 2^k$ Walsh matrix. The Hadamard matrix $H_k$ for WHT of $X$ is constructed as follows:
\begin{equation}
H_k = \left\{
        \begin{array}{ll}
            1, & k=0\\
           \begin{bmatrix}
            $$H_{k-1}$$ & $$H_{k-1}$$\\
            $$H_{k-1}$$ & $$-H_{k-1}$$
        \end{bmatrix}, & k>0
        \end{array}
    \right.
\end{equation}
The Hadamard matrix is then rearranged to increase the sign change order, resulting in the Walsh matrix. This matrix exhibits the unique property of orthogonality, where every row of the matrix is orthogonal to each other, with the dot product of any two rows being zero. This property makes the Walsh matrix particularly advantageous in a wide range of signal and image processing applications \cite{jayathilake2013discrete}.

However, WHT poses a computational challenge when the dimension of the input vector is not a power of two. To address this, a technique called blockwise Walsh-Hadamard transforms (BWHT) was introduced \cite{pan2022block}. BWHT divides the transform matrix into multiple blocks, each sized to an integer power of two, significantly reducing the worst-case size of operating tensors and mitigating excessive zero-padding.

\subsection{Frequency-Domain Compression of Deep Neural Networks}
Frequency-domain transformations, such as BWHT, can be incorporated into deep learning architectures for model compression. For instance, in MobileNetV2, which uses $1 \times 1$ convolutions in its bottleneck layers to reduce computational complexity, BWHT can replace these convolution layers, achieving similar accuracy with fewer parameters. Unlike $1 \times 1$ convolution layers, BWHT-based binary layers use fixed Walsh-Hadamard matrices, eliminating trainable parameters. Instead, a soft-thresholding activation function with a trainable parameter $T$ can be used to selectively attend to important frequency bands. The activation function $S_T$ for this frequency-domain compression is given as
\begin{equation}
y = S_{T}(x) = sign(x)(|x| - T) = \left\{
                                        \begin{array}{ll}
                                            x+T, & x < -T\\
                                            0, & |x| \le T\\
                                            x-T, & x > T
                                        \end{array}
                                    \right.
\end{equation}
Similarly, in ResNet20, $1 \times 1$ convolutions can be replaced with BWHT layers. These transformations maintain matching accuracy while achieving significant compression on benchmark datasets such as CIFAR-10, CIFAR-100, and ImageNet \cite{pan2022block}. However, BWHT transforms also increase the necessary computations for deep networks. The subsequent section discusses how micro-architectures and circuits can enhance the computational efficiency of BWHT-based tensor transformations.

\begin{figure}[t!]
\centering
    \includegraphics[width=0.9\linewidth]{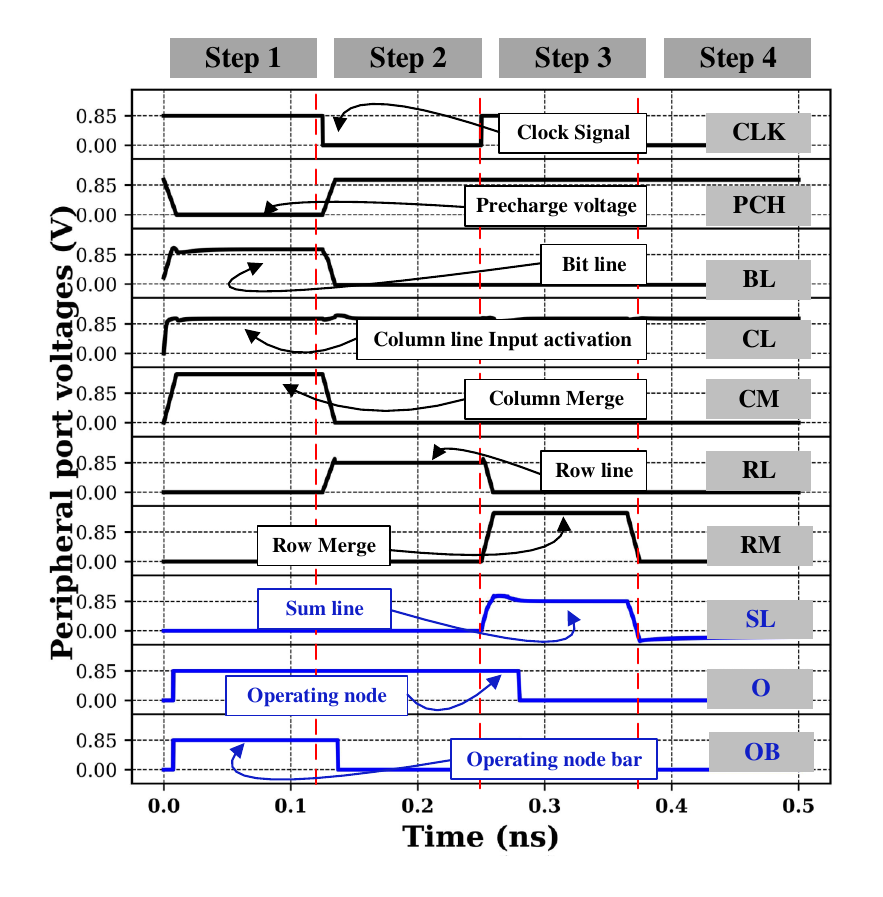}   
    \caption{\textbf{Timing diagram of signal flows:} Waveforms of key signals in the CIM operation, including clock signal (CLK), precharge signal (PCH), bit lines (BL/BLB), column lines (CL/CLB), column-merge signal (CM), row lines (RL), row-merge signal (RM), sum lines (SL/SLB), and Operating points (O/OB). The four-step CIM operation is completed in just two clock cycles (\textbf{4 GHz}), accelerating processing times and enhancing performance. The compact cell design, comprised solely of NMOS transistors, ensures efficient use of space. Boosting techniques are applied to CM and RM signals to eliminate the impact of threshold voltage, further optimizing the operation of our CIM design.}
    \label{fig: waveforms}
\end{figure}

\section{Analog-Domain Frequency Transforms}

The escalating data deluge in the digital world has brought analog domain processing to the forefront as a potential solution for accelerating vector and matrix-level parallel instructions, particularly for low-precision computations such as matrix-vector products in DNNs. Analog computations, by their nature, simplify processing cells and allow for the integration of storage and computation within a single cell. This feature is prevalent in many compute-in-memory designs. This integration significantly mitigates the data deluge by reducing data movement during deep learning computations, a critical bottleneck for energy and performance in traditional processors \cite{9987520}.

However, analog domain processing does not resolve the data deluge challenge entirely due to its reliance on ADCs and DACs for domain conversions of operands. The ADC and DAC operations introduce design complexities, significant area/power overheads, and limitations in technology scalability, all contributing to the data deluge. Furthermore, the performance of ADCs and DACs is constrained by speed, power consumption, and cost, further limiting the overall capabilities of analog domain computations. In the subsequent discussion, we present our proposed techniques for analog domain processing of frequency operations. These aim to alleviate the data deluge by eliminating the need for ADC/DAC conversions, even when operating on digital input vectors and computing output vectors in the digital domain. Our approach incorporates bitplane-wise input vector processing and co-designs learning methods that can operate accurately under extreme quantization, enabling ADC/DAC-free operations.

\subsection{Crossbar Microarchitecture Design and Operation}
In Fig. \ref{fig:top-level architecture}, we propose a design that leverages analog computations for frequency domain processing of neural networks to address the data deluge. The design's crossbar combines six transistors (6T) NMOS-based cells for analog-domain frequency transformation. The corresponding cells for `-1' and `1' entries in the Walsh-Hadamard transform matrix are shown to the figure's right. The crossbar combines these cells according to the elements in the transform matrix. Since the transform matrix is parameter-free, computing cells in the proposed design are simpler by being based only on NMOS for a lower area than conventional 6T or 8T SRAM-based compute-in-memory designs. Additionally, processing cells in the proposed crossbar are \textit{stitchable} along rows and columns to enable perfect parallelism and extreme throughput, mitigating the data deluge further. 

The operation of the crossbar comprises four steps, which are marked in Fig. \ref{fig:top-level architecture}. These steps are designed to minimize data movement and thus address the data deluge. The steps include precharging, independent local computations, row-wise summing, and single-bit output generation.

\begin{figure}[t!]
    \centering \includegraphics[width=0.99\linewidth]{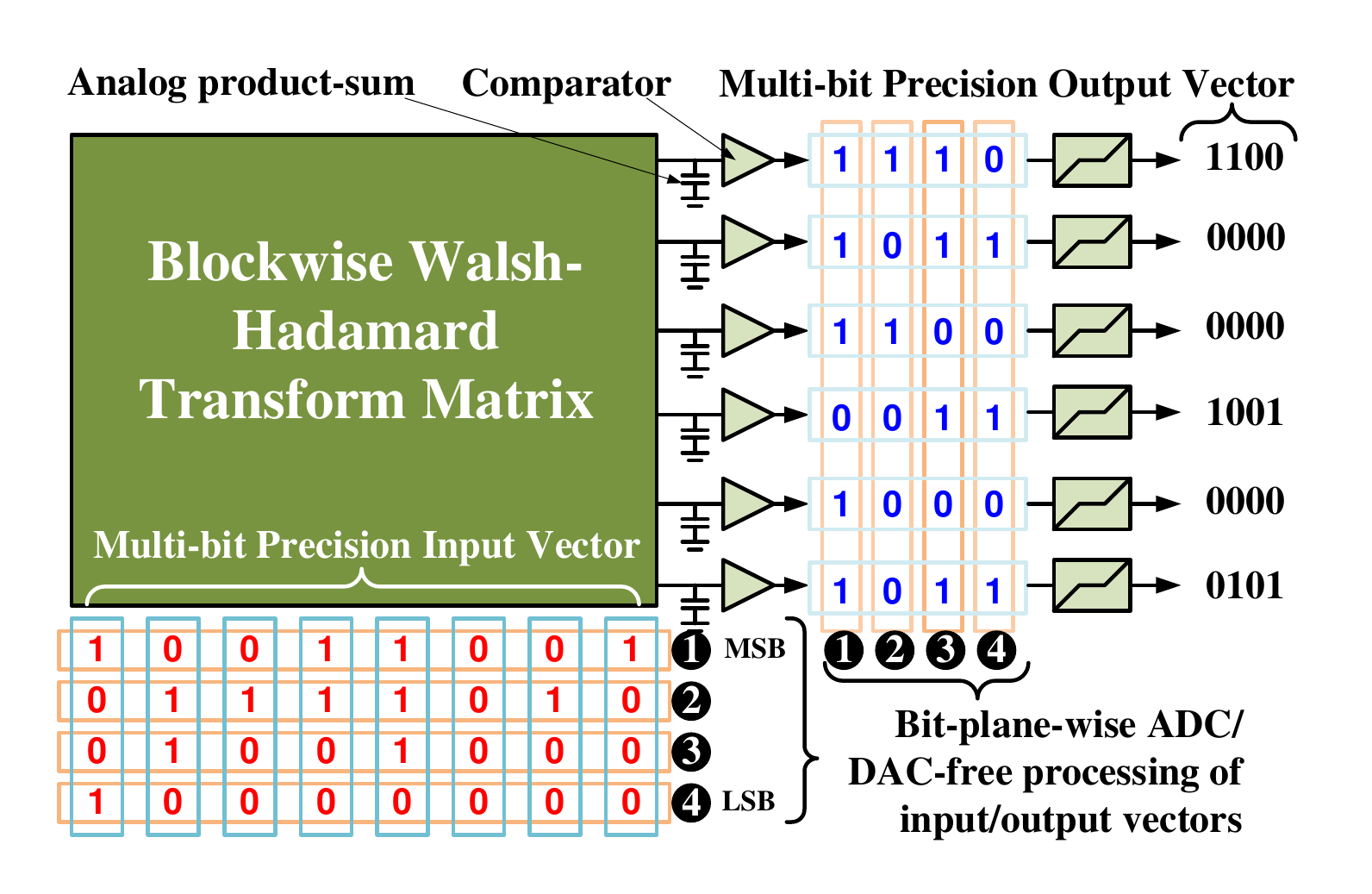}
    \caption{\textbf{Bitplane-wise operation flow:} A multibit input vector is processed in a bitplane-wise manner. The same significance bits of vector elements are grouped together and processed in a single step. Analog the crossbar rows output vector bits are computed in parallel. The output bits generated for all input bitplanes are concatenated to produce the multibit output vector.}
    \label{fig:operation-flow}
\end{figure}

Fig. \ref{fig: waveforms} shows the signal flow diagram for the above four steps. The 16 nm predictive technology models (PTM) simulation results are shown using the low standby power (LSTP) library in \cite{PTM} and operating the system at 4 GHz clock frequency and VDD = 0.85V. Row-merge and column-merge signals in the design are boosted at 1.25V to avoid source degeneration. Unlike comparable compute-in-memory designs such as \cite{9972346}, which place product computations on bit lines, in our design, these computations are placed on local nodes in parallel at all array cells. This improves parallelism, energy efficiency, and performance by computing on significantly less capacitive local nodes than bit lines in traditional designs, thereby further addressing the data deluge.     

\subsection{ADC-Free by Training against 1-bit Quantization}

Fig. \ref{fig:operation-flow} presents the high-level operation flow of our scheme for processing multi-bit digital input vectors and generating the corresponding multi-bit digital output vector using the analog crossbar illustrated in Fig. \ref{fig:top-level architecture}. This scheme is designed to address the data deluge by eliminating the need for ADC and DAC operations. The scheme utilizes bitplane-wise processing of the multi-bit digital input vector and is trained to operate effectively with extreme quantization. In the figure, the input vector's elements with the same significance bits are grouped and processed in a single step using the scheme described in Fig. \ref{fig:top-level architecture}, which spans two clock cycles. The analog charge-represented output is computed along the row-wise charge sum lines and thresholded to generate the corresponding digital bits. This extreme quantization approach is applied to the computed MAC output, eliminating the need for ADCs. With multiple input bitplanes, labeled as \ballnumber{1}-\ballnumber{2}-\ballnumber{3}-\ballnumber{4} in the figure, the corresponding output bitplanes are concatenated to form the final multi-bit output vector.

\begin{figure}[t!]
    \centering  
    \includegraphics[width=0.47\linewidth]{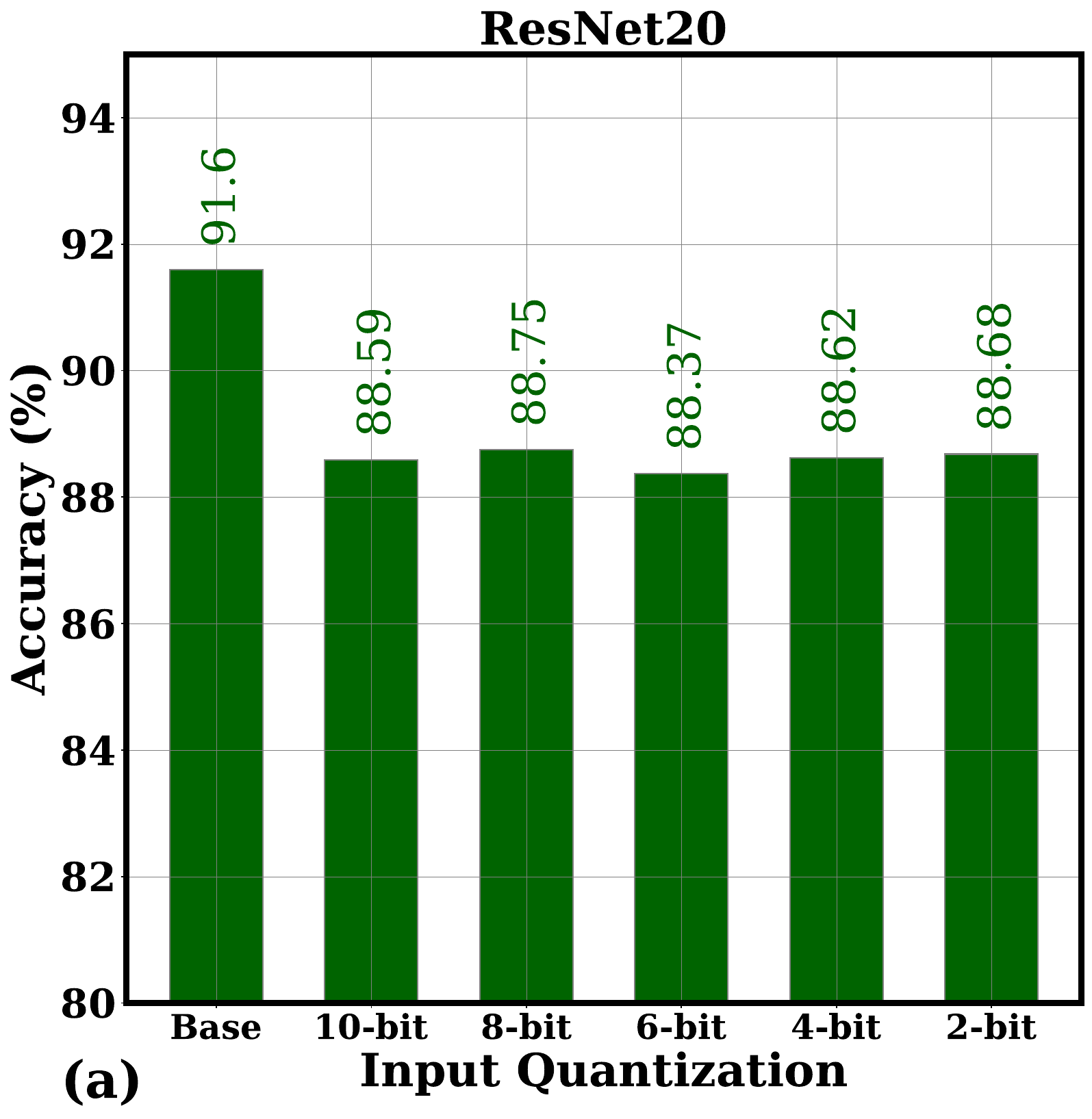}
    \includegraphics[width=0.49\linewidth]{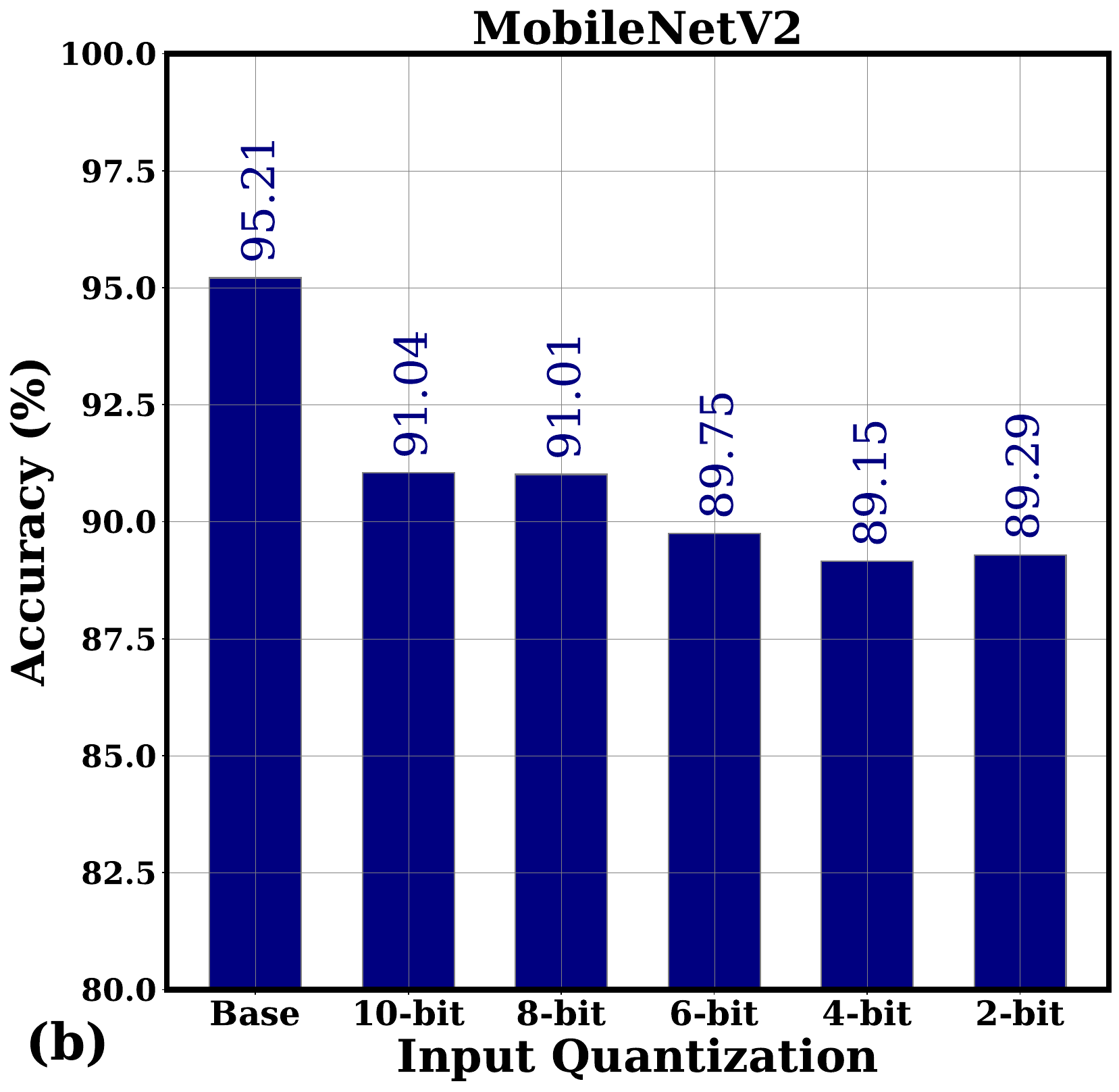}  
    \caption{\textbf{Accuracy under training with 1-bit quantization:} Impact of input quantization on the performance of deep learning models \textbf{(a)} ResNet20 and \textbf{(b)} MobileNetV2 on the CIFAR-10 dataset, while considering 1-bit product-sum quantization and varying input quantization levels. The results demonstrate that accuracy converges to a similar level across all input quantization levels, and it is 3-4\% lower than the floating-point baseline. }
    \label{fig:Accuracy results for MobileNet and ResNet}
\end{figure}

\begin{figure*}[h!]
    \centering
    \includegraphics[width=0.32\linewidth]{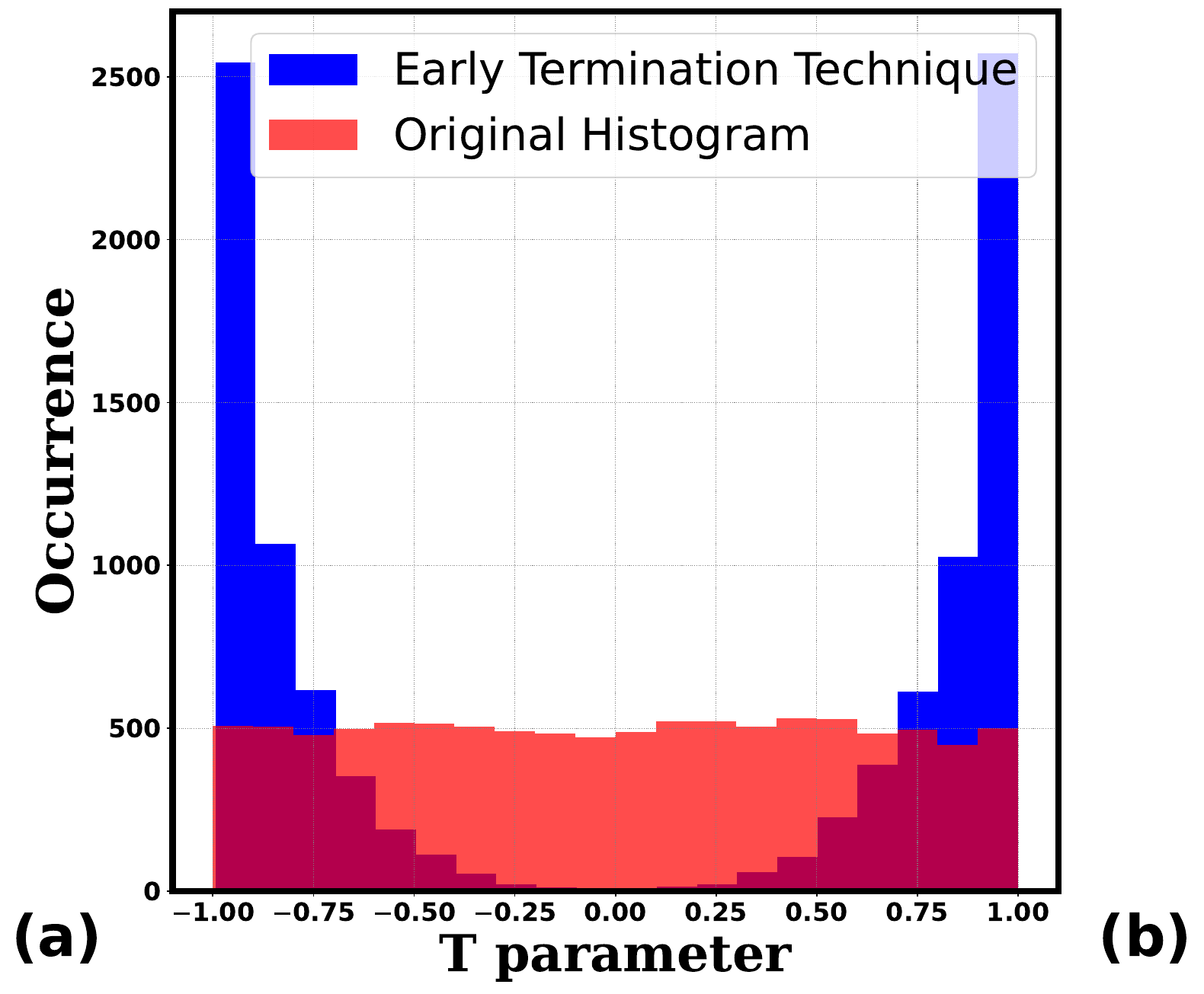}
    \includegraphics[width=0.32\linewidth]{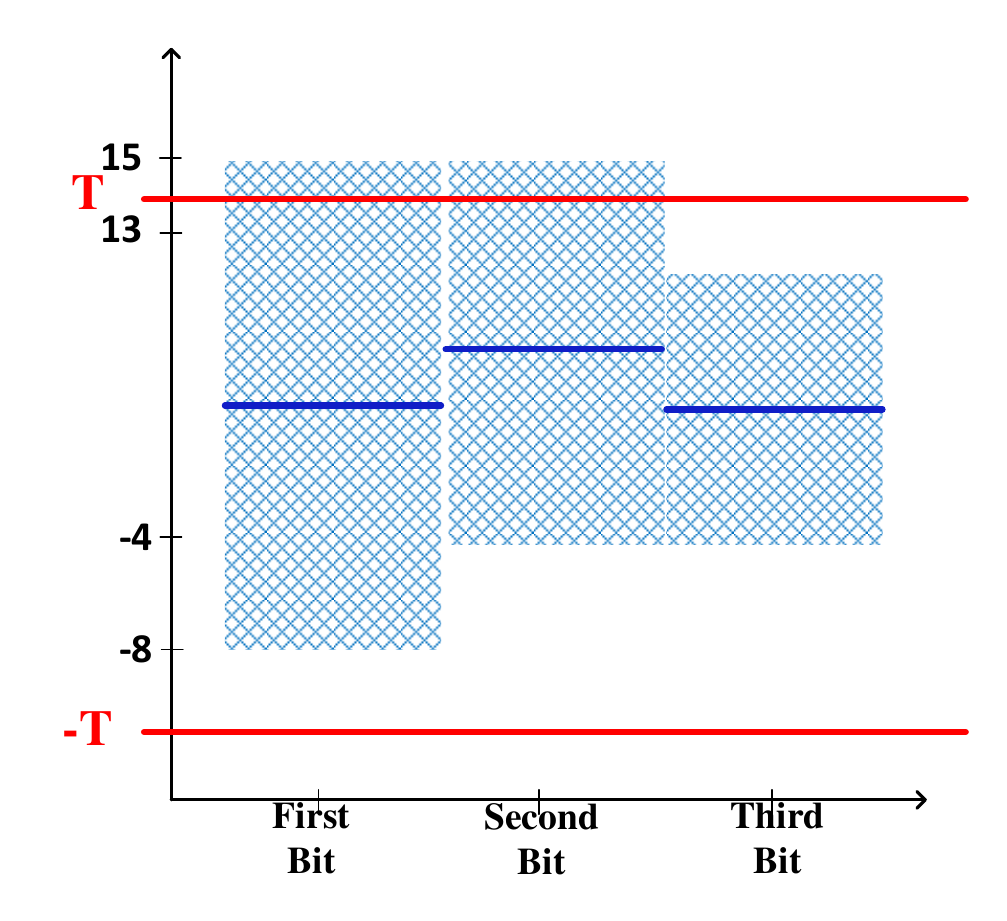}
    \includegraphics[width=0.27\linewidth]{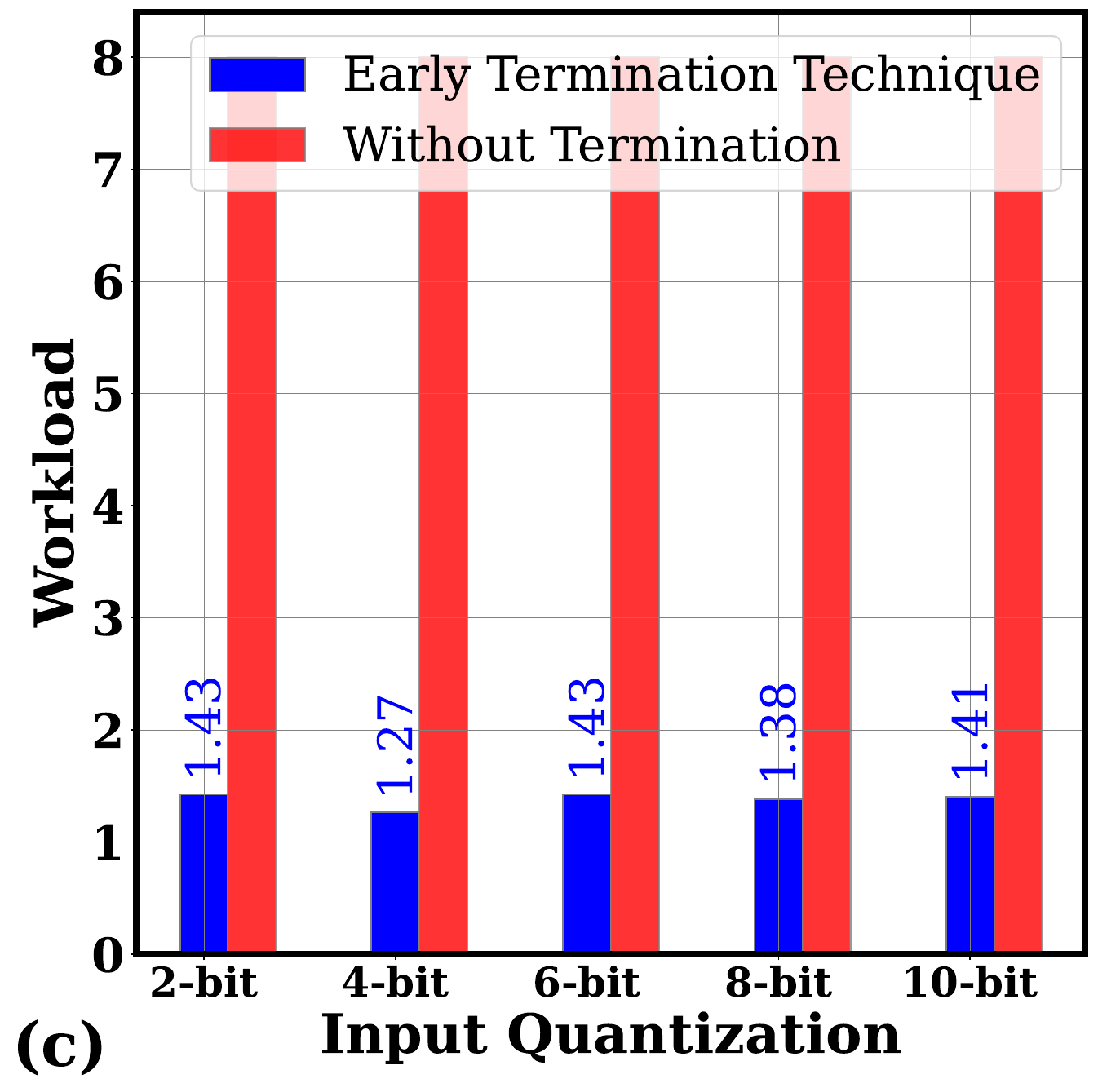}
    \caption{ \textbf{Early Termination Technique:} This figure illustrates the impact of the early termination technique on the distribution of the soft-thresholding parameter (T). It shows how applying a unique loss function drives the T parameter towards (-1) and (1), aiding in workload reduction. The figure also presents a scenario where the value, after processing through three bit-planes, falls within the range of (-T) and (T), indicating a zero output and eliminating the need for further processing of the remaining input bits. The effectiveness of early termination techniques in reducing workload, maintaining accuracy, and their independence from the level of weight quantization are also demonstrated.}
    \label{fig:Early Termination Process}
\end{figure*}

\begin{figure*}[h!]
    \centering
    \includegraphics[width=\linewidth]{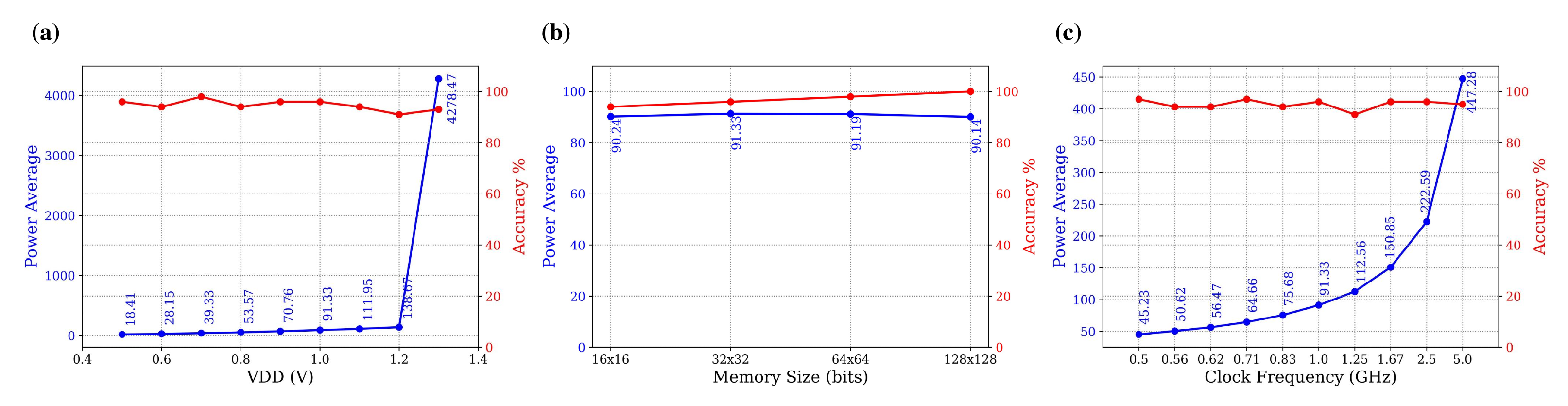}   
    \caption{ \textbf{Performance analysis of proposed CIM architecture:} \textbf{(a)} Examination of the supply voltage (VDD) impact on power consumption and accuracy, emphasizing the marked increase in power consumption at 1.3 volts. In this case, the clock frequency is 1 (GHz), and the memory size is $32\times32$ (bits). \textbf{(b)} Evaluation of the CIM architecture's accuracy and power consumption across different memory array sizes ($16\times16, 32\times32, 64\times64,$ and $128\times128$ (bits)), demonstrating the persistently high accuracy attributable to the highly parallel design. The supply voltage and clock frequency are set at 1 (V) and 1 (GHz). \textbf{(c)} Investigation of power consumption and accuracy trends concerning clock frequency, revealing that beyond 2.5 (GHz), the average power consumption escalates significantly, thus restricting the overall performance of the circuit. The supply voltage and memory size are 1 (V) and $32\times32$ (bits).}
    \label{fig:Hspice}
\end{figure*}

Due to the extreme quantization applied to the analog output in the above scheme, the resulting digital output vector only approximates the true output vector under transformation. However, unlike a standard DNN weight matrix, the transformation matrix used in our frequency domain processing is parameter-free. This characteristic enables training the system to effectively mitigate the impact of the approximation while allowing for a significantly simpler implementation without needing ADCs or DACs, thereby further addressing the data deluge. The following training methodology achieves this.

Consider the frequency-domain processing of an input vector $\vectorbold{x_i}$. In Fig. \ref{fig:operation-flow}, we process $\vectorbold{x_i}$ by transforming it to the frequency domain, followed by parameterized thresholding, and then reverting the output to the spatial domain. Consider a DNN with $n$ layers that chain the above sequence of operations as $\vectorbold{x_{i+1}} = F_0(S_{T,i}(F_0(\vectorbold{x_{i}}))) $. Here, $F_0()$ is a parameter-free \textit{approximate} frequency transformation as followed in our scheme in Fig. \ref{fig:operation-flow}. $S_{T,i}()$ is a parameterized thresholding function at the i$^{th}$ layer whose parameters $T_i$ are learned from the training data. 

\begin{figure*}[t!]
  \centering
  \includegraphics[width=0.58\linewidth]{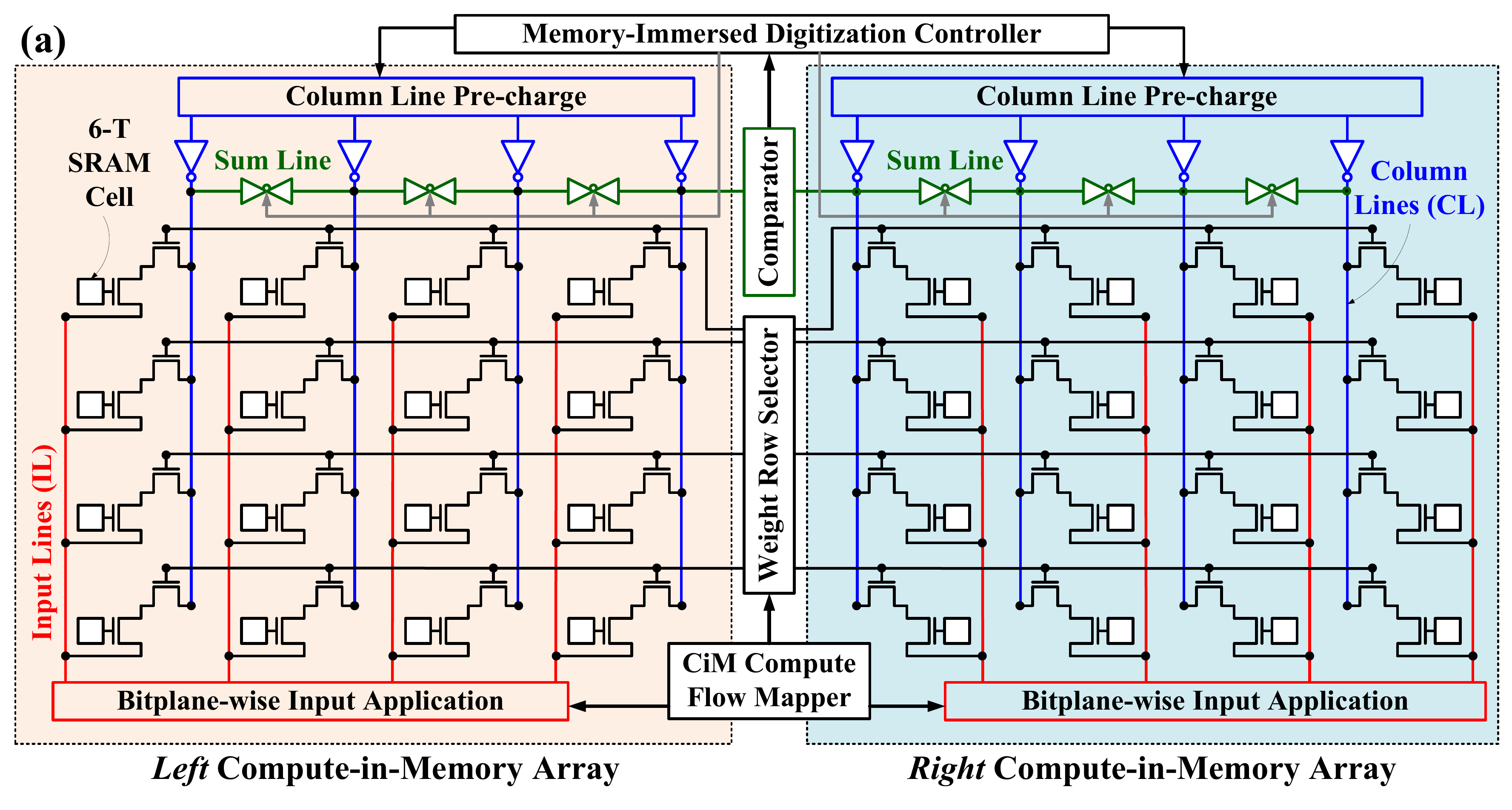}
  \includegraphics[width=0.4\linewidth]{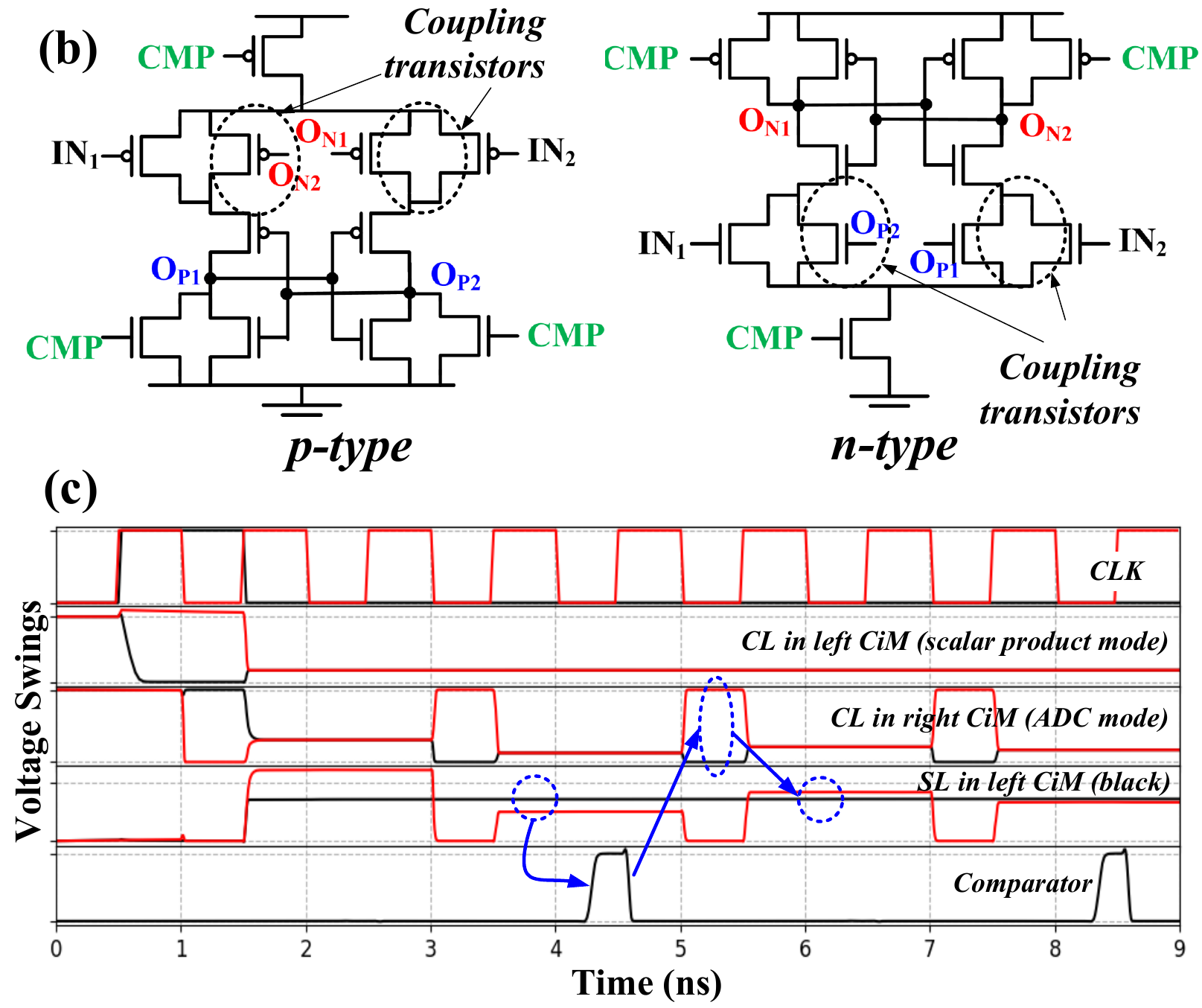}
  \caption{\textbf{Architecture and waveforms of SRAM-immersed ADC:} \textbf{(a)} Coupling of left-right memory arrays for memory-immersed digitization. When the left array computes within-memory scalar product, the right array digitizes analog-domain computed output. Both arrays switch their operation subsequently for collaborative digitization. \textbf{(b)} Clocked comparator design combining n-type and p-type counterparts for rail-to-rail voltage comparison. \textbf{(c)} Transient waveforms.}
  \label{fig:arch}
\end{figure*}

\subsection{Early Termination for Energy Efficiency}

To further mitigate the data deluge, we introduce an early termination mechanism in our design. This mechanism is based on the observation that the contribution of higher bitplanes in the input vector to the final output is significantly less than the lower bitplanes due to the binary representation of numbers. Therefore, we can terminate the computation early if the higher bitplanes do not significantly affect the final output, thus saving energy and reducing data movement.

We illustrate the early termination mechanism in Fig. \ref{fig:Early Termination Process}. The mechanism involves a thresholding operation that compares the partial sum of the output bitplanes with a predefined threshold. If the partial sum is less than the threshold, the computation is terminated early, and the remaining higher bitplanes are ignored. This mechanism reduces the number of computations and data movement, thereby mitigating the data deluge. The threshold for early termination is a design parameter that can be tuned based on the accuracy-energy trade-off. A lower threshold leads to more frequent early terminations, saving more energy but potentially reducing the accuracy, and vice versa. Therefore, the threshold should be carefully chosen to balance the trade-off between energy efficiency and accuracy.

Our proposed techniques for analog domain processing of frequency operations, including bitplane-wise input vector processing, ADC/DAC-free operations, and early termination mechanism, effectively address the data deluge by reducing data movement and computational complexity. These techniques also maintain high accuracy and energy efficiency, making them promising for future low-precision computations in deep neural networks. Fig. \ref{fig:Hspice} shows our simulation results supporting the claim.

\begin{figure}[h!]
  \centering
  \includegraphics[width=0.8\linewidth]{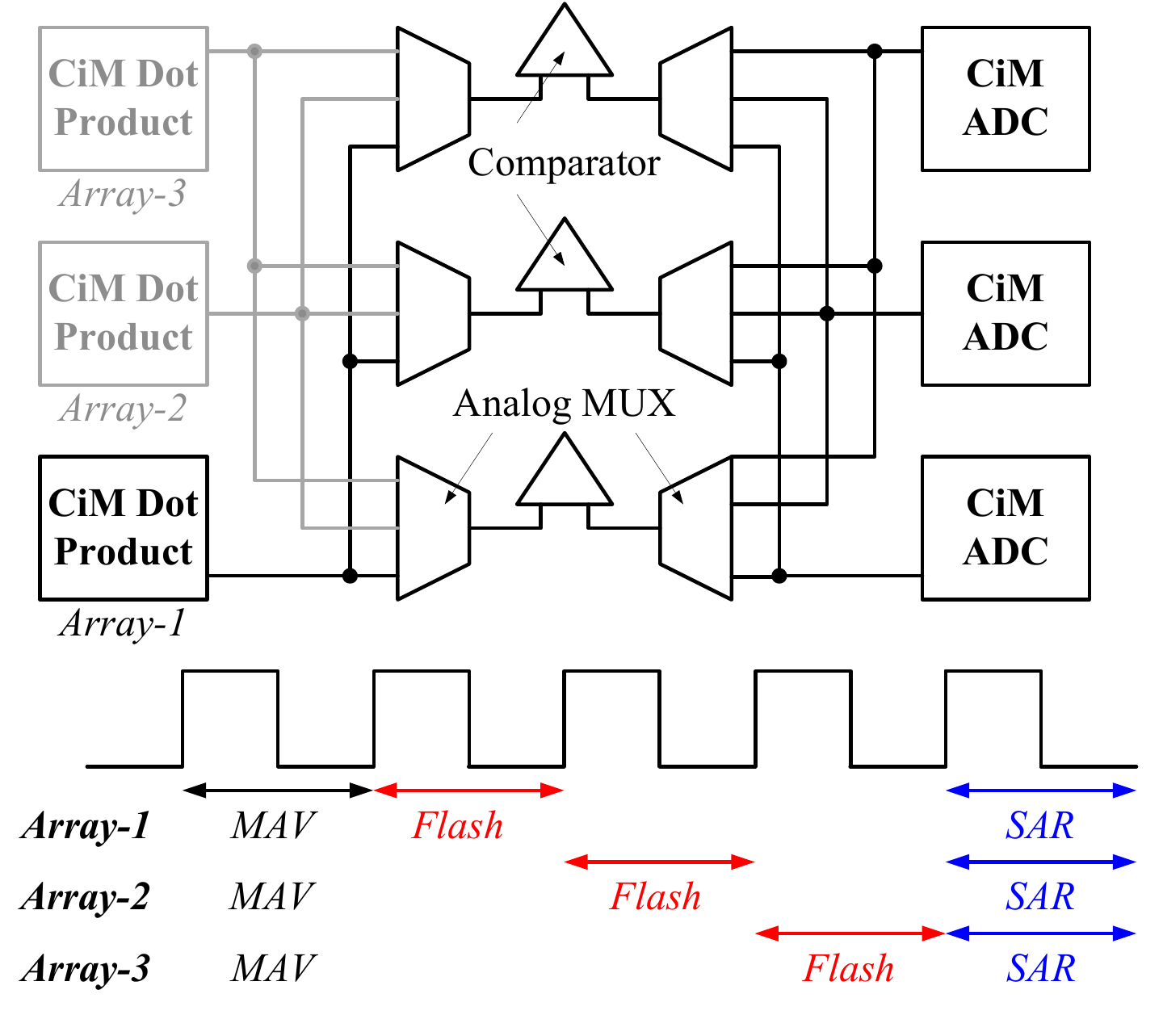}
  \caption{\textbf{Hybrid mode of SRAM-immersed ADC:} A dot product-configured CiM array is coupled to many ADC-configured arrays to the right for flash mode digitization of the initial most significant bits. After this, each left-array couples to the nearest right array to determine the remaining bits in SAR mode. Operational cycles are shown at the bottom.}
  \label{fig:chip}
\end{figure}
\section{Memory-Immersed Collaborative Digitization}
\subsection{Integrating CiM Arrays for Collaborative Digitization}
Fig. \ref{fig:Introduction}(b) illustrates the implementation of memory-integrated ADC. We specifically focus on our methods and results for eight-transistor (8T) compute-in-SRAM arrays, widely utilized in numerous platforms. Unlike 6T compute-in-SRAM, 8T cells are less prone to bit disturbances due to separate write and inference ports, making them more suitable for technology scaling. However, our suggestions for memory-integrated ADC apply to other memory types, including 10-T compute-in-SRAM/eDRAM and non-volatile memory crossbars.

In the proposed model, two adjacent CiM arrays work together for in-memory digitization, as depicted in Figure 8(a). When the left array calculates the input-weight scalar product, the right array performs SRAM-integrated digitization on the produced analog-mode multiply-average (MAV) outputs. Both arrays then switch their operating modes. Each array consists of 8-T cells, combining standard 6-T SRAM cells with a two-transistor weight-input product port shown in the figure. Memory cells for input-weight products are accessed using three control signals. 

For in-memory digitization of charge-domain product-sum computed in the left array, column lines (CLs) in the right array realize the unit capacitors of a capacitive DAC formed within the memory array. A precharge transistor array is integrated with the column lines to generate the reference voltages. The first reference voltage is generated by summing the charges of all column lines. The developed MAV voltage in the left CiM array is compared to the first reference voltage to determine the most significant bit of the digitized output. The next precharge state of memory-immersed capacitive DAC is determined, and the precharge and comparison cycles continue until the MAV voltage has been digitized to necessary precision.

Using neighboring CiM arrays for the first reference voltage generation for in-memory digitization offers several key advantages. First, various non-idealities in analog-mode MAV computation become common-mode due to using an identical array for the first reference voltage generation. Thus, the non-idealities only minimally impact the accuracy of digitization. Second, collaborative digitization minimizes peripheral overheads. Only an analog comparator and simple modification in the precharge array are sufficient to realize a successive approximation search. 

Compared to traditional CiM approaches with a dedicated ADC at each array, our scheme's interleaving of scalar product computation and digitization cycles affects the achievable throughput. However, with simplified low-area peripherals, more CiM arrays can be accommodated than prior works employing dedicated ADCs. Therefore, our scheme compensates for the overall throughput at the system level by operating many parallel CiM arrays. This improved area efficiency of CiM arrays in our scheme minimizes the necessary exchanges from off-chip DRAMs to on-chip structures in mapping large DNN layers, a significant energy overhead in conventional techniques.

\begin{figure}[t!]
  \centering
  \includegraphics[width=\linewidth]{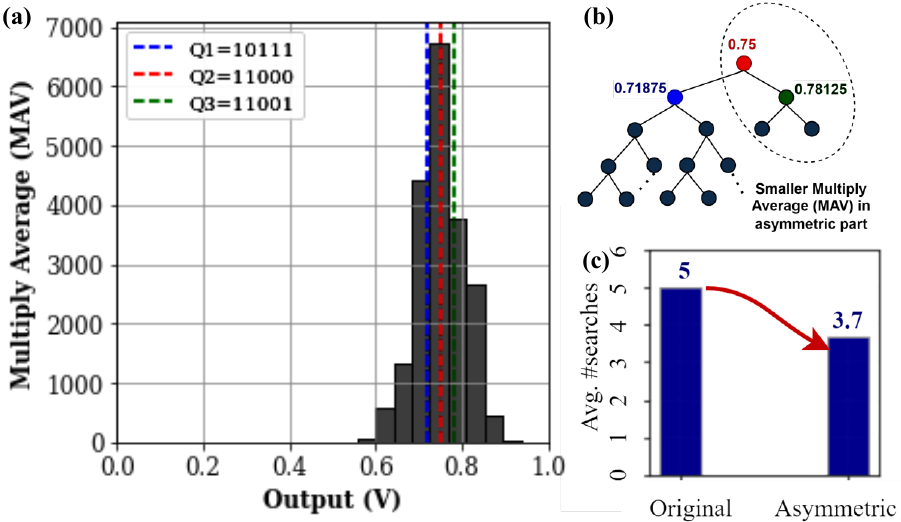}
  \caption{\textbf{Exploiting MAV statistics for ADC's time-efficiency:} \textbf{(a)} Distribution of MAV under the uniform distribution of input and weight bits for CiM scheme in Figure 2. \textbf{(b)} Asymmetric binary search for skewed MAV statistics. \textbf{(c)} For 5-bit data conversion, asymmetric search requires on average $\sim$3.7 comparisons, unlike symmetric binary search that requires $\sim$5 comparisons.}
  \label{fig:statistics}
\end{figure}

\subsection{Hybrid SRAM-Integrated Flash and SAR ADC Operation}
In addition to the nearest neighbor networking in Figure 8, more complex CiM networks can also be orchestrated for more time-efficient collaborative digitization in Flash and/or hybrid SAR + Flash mode. Figure 9 shows an example networking scheme where Array-1 couples with three memory arrays to the right for collaborative digitization in Flash mode. Here, the three right arrays simultaneously generate the respective reference voltages for the Flash mode of digitization and to determine the first two most significant bits in one comparison cycle.

\begin{figure*}[t!]
  \centering
  \includegraphics[width=\linewidth]{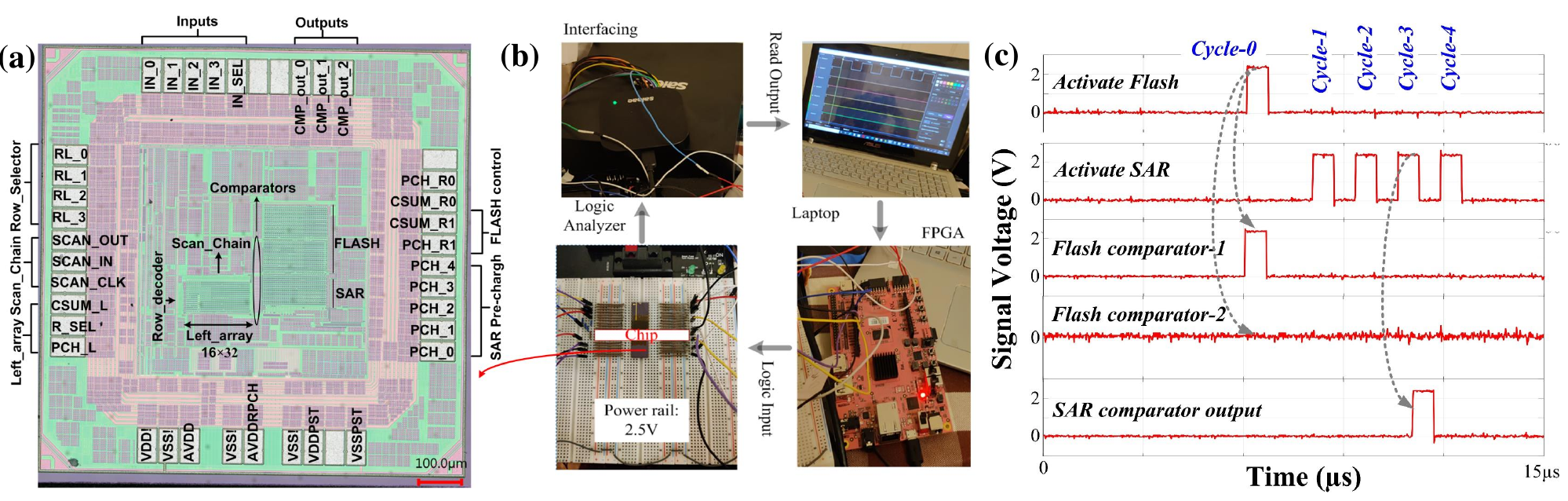}
  \caption{\textbf{Test-chip and measurements:} \textbf{(a)} Micrograph of fabricated design in 65 nm CMOS. Four compute-in-SRAM arrays, A1$-$A4 were fabricated. A1 interfaces with A2 to realize SRAM-immersed SAR ADC. A1 interfaces with A1$-$A4 to realize SRAM-immersed Flash ADC. \textbf{(b)} Measurement setup. \textbf{(c)} Measurement transient waveforms for hybrid SAR + Flash ADC operation.}
  \label{fig:chip}
\end{figure*}

\subsection{Leveraging MAV Statistics for ADC's Time-Efficiency}
The hybrid scheme for data conversion further benefits from exploiting the statistics of the multiply average (MAV) computed by the CiM. In many CiM schemes, the computed MAV is not necessarily uniformly distributed. For example, in bit-plane-wise CiM processing, DACs are avoided by processing a one-bit input plane in one time step. This results in a skewed distribution of MAV, Fig. 10(a). The skewed distribution of MAV can be leveraged by implementing an asymmetric binary search for digitization, Fig. 10(b). Under the asymmetric search, the average number of comparisons reduces, thus proportionally reducing the energy and latency for the operation, Fig. 10(c). The proposed hybrid digitization scheme further exploits the asymmetric search, which can be accelerated by Flash digitization mode.

\subsection{Test-Chip Design, Measurement Results, and Comparison to Traditional ADC}
A 65 nm CMOS test chip characterized the proposed SRAM-integrated ADC. The fabricated chip's micrograph and measurement setup are shown in Figures 11(a, b). Four compute-in-SRAM arrays of size 16x32 were implemented. The coupling of CiM arrays can also be programmed to realize hybrid Flash-SAR ADC operations, such as obtaining the two most significant bits in Flash mode and the remaining in SAR.

\begin{figure*}[t!]
  \centering
  \includegraphics[width=0.85\linewidth]{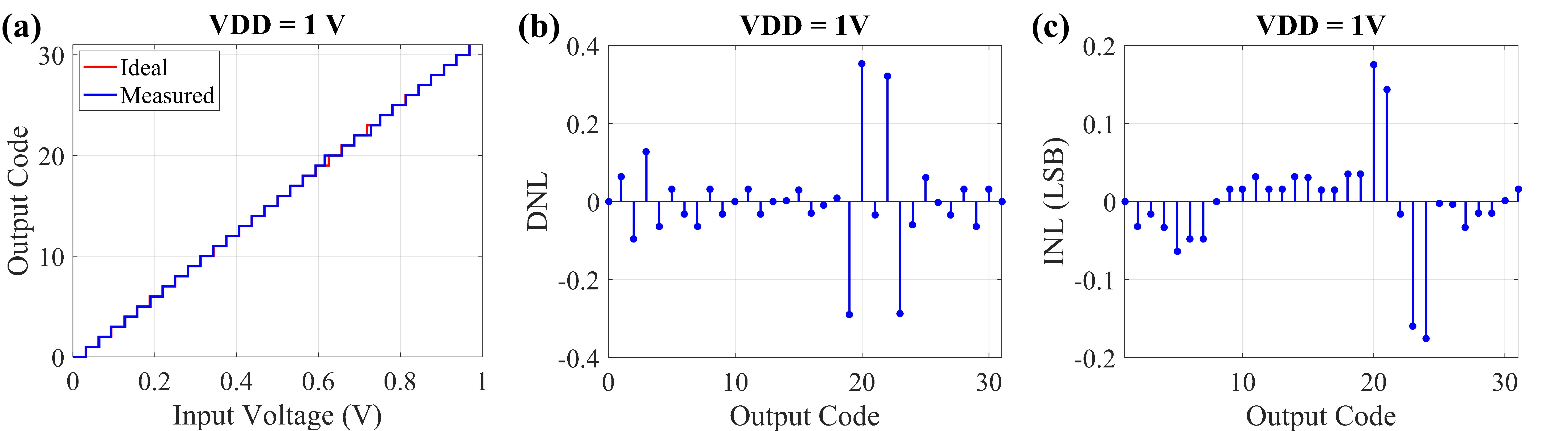}
  \caption{\textbf{Measured non-idealities of SRAM-immersed ADC:} (a) Output code vs. applied input voltage. (b) Differential and (c) integrated non-linearities.}
  \label{fig:INL_DNL}
\end{figure*}

\begin{center}
\textbf{Table I: Comparison of 5-bit in-memory ADC with 10 MHz clock against SAR and Flash architectures} \\ 
\vspace{5pt}
\begin{tabular}{ p{3cm} p{1.3cm} p{1.3cm} p{1.3cm}} 
\hline
\textbf{Architecture }& Tech. & Area ($\mu$m$^2$) & Energy (pJ) \\ \hline \hline
\textbf{SAR} \cite{jiang2021analog} & 40 nm & 5235.20 & 105 \\ \hline
\textbf{Flash} \cite{jiang2021analog} & 40 nm & 10703.36 & 952 \\ \hline
\textbf{In-Memory} (ours) & 65 nm & 207.8 & 74.23 \\ \hline
\end{tabular}
\end{center}

Figure 11(c) shows the transient waveforms of different control signals and comparator outputs, showing a hybrid Flash + SAR ADC operation. Flash mode is activated in the first comparison cycle where CiM arrays generate the corresponding reference voltages, and the first two bits of MAV digitization are extracted. Subsequently, the operation switches to SAR mode, where the remaining digitization bits are obtained by engaging one array alone with another. In the last four cycles, other arrays become free to similarly operate on a proximal CiM array to digitize MAV in SAR mode. Figure 12(a) shows the measured staircase plot of input voltage to output codes and the comparison to an ideal staircase, demonstrating near-ideal performance. This suggests that the proposed SRAM-integrated ADC operates effectively, with the hybrid Flash + SAR ADC operation providing a flexible and efficient approach to digitization within the memory array.

Fig. 13 and Table I show the design space exploration of pro-
posed memory-immersed ADC compared to other ADC styles.
In Fig. 13(a), leveraging in-memory structures for capacitive
DAC formation, the proposed in-memory ADC is more area
efficient than Flash and SAR styles. Significantly, Flash ADC’s
size increases exponentially with increasing bit precision. In
Fig. 13(b), SAR ADC’s latency increases with bit precision
while Flash ADC can maintain a consistent latency but at the
cost of the increasing area as shown in Fig. 13(a). A hybrid
data conversion in the proposed in-memory provides a middle
ground, i.e., lower latency than in SAR ADC. Figs. 13(c-d)
show the impact of supply voltage and frequency scaling on
in-memory ADC’s power and accuracy for MNIST character
recognition.

In conclusion, the proposed method of integrating compute-in-memory (CiM) arrays for collaborative digitization offers a promising approach to handling the data deluge in modern computing systems. By leveraging the unique properties of 8T compute-in-SRAM arrays, the method provides a scalable and efficient solution for in-memory digitization. The hybrid SRAM-integrated Flash and SAR ADC operation further enhances time efficiency, while the exploitation of MAV statistics contributes to the ADC's time efficiency. The successful implementation and characterization of the proposed SRAM-integrated ADC on a 65 nm CMOS test chip further validate the effectiveness of this approach.

\section{Sustainability}
The proposed CIM architecture and the associated techniques significantly contribute to sustainability in deep learning applications. The key to this sustainability lies in the efficient use of resources and the reduction of energy consumption.

Firstly, the architecture leverages BWHT and soft-thresholding techniques to compress deep neural networks. This compression reduces the computational resources required, leading to more efficient use of hardware. By reducing the number of parameters in the BWHT layer, the architecture minimizes the memory footprint of deep learning models, reducing the energy required for data storage and retrieval. The early termination strategy enhances energy efficiency by leveraging output sparsity to reduce computation time. 

Secondly, the memory-immersed collaborative digitization among CiM arrays minimizes the area overheads of ADCs for deep learning inference. This allows significantly more CiM arrays to be accommodated within limited footprint designs, improving parallelism and minimizing external memory accesses. The results demonstrate the potential of the proposed techniques for area-efficient and energy-efficient deep learning applications.

Lastly, the proposed techniques and architectures are designed to be robust and resilient, reducing the need for frequent hardware replacements or upgrades. This longevity contributes to sustainability by reducing electronic waste and the environmental impact associated with the production and disposal of hardware.

\section{Conclusions}

The proposed frequency-domain CIM architecture with early termination technique, and memory-immersed collaborative digitization present a comprehensive solution for sustainable and efficient deep learning applications. This leads to more efficient use of hardware and reduces the energy required for data storage and retrieval. The memory-immersed collaborative digitization among CiM arrays minimizes the area overheads of a conventional ADC for deep learning inference. This allows significantly more CiM arrays to be accommodated within limited footprint designs, improving parallelism and minimizing external memory accesses. The results demonstrate the potential of the proposed techniques for area-efficient and energy-efficient deep learning applications.

These techniques contribute significantly to sustainability in deep learning applications. By improving area efficiency in deep learning inference tasks and reducing energy consumption, these techniques contribute to the sustainability of data processing at the edge. This approach enables better handling of high-dimensional, multispectral analog data. It helps alleviate the challenges the analog data deluge poses, making it a promising solution for sustainable data processing at the edge. In conclusion, the proposed techniques and architectures pave the way for the next generation of deep learning applications, particularly in scenarios where area and power resources are limited.

\section{Acknowledgment}
This work was supported by COGNISENSE, one of the seven centers in JUMP 2.0, a Semiconductor Research Corporation (SRC) program sponsored by DARPA. 

\begin{figure}[H]
  \centering
  \includegraphics[width=\linewidth]{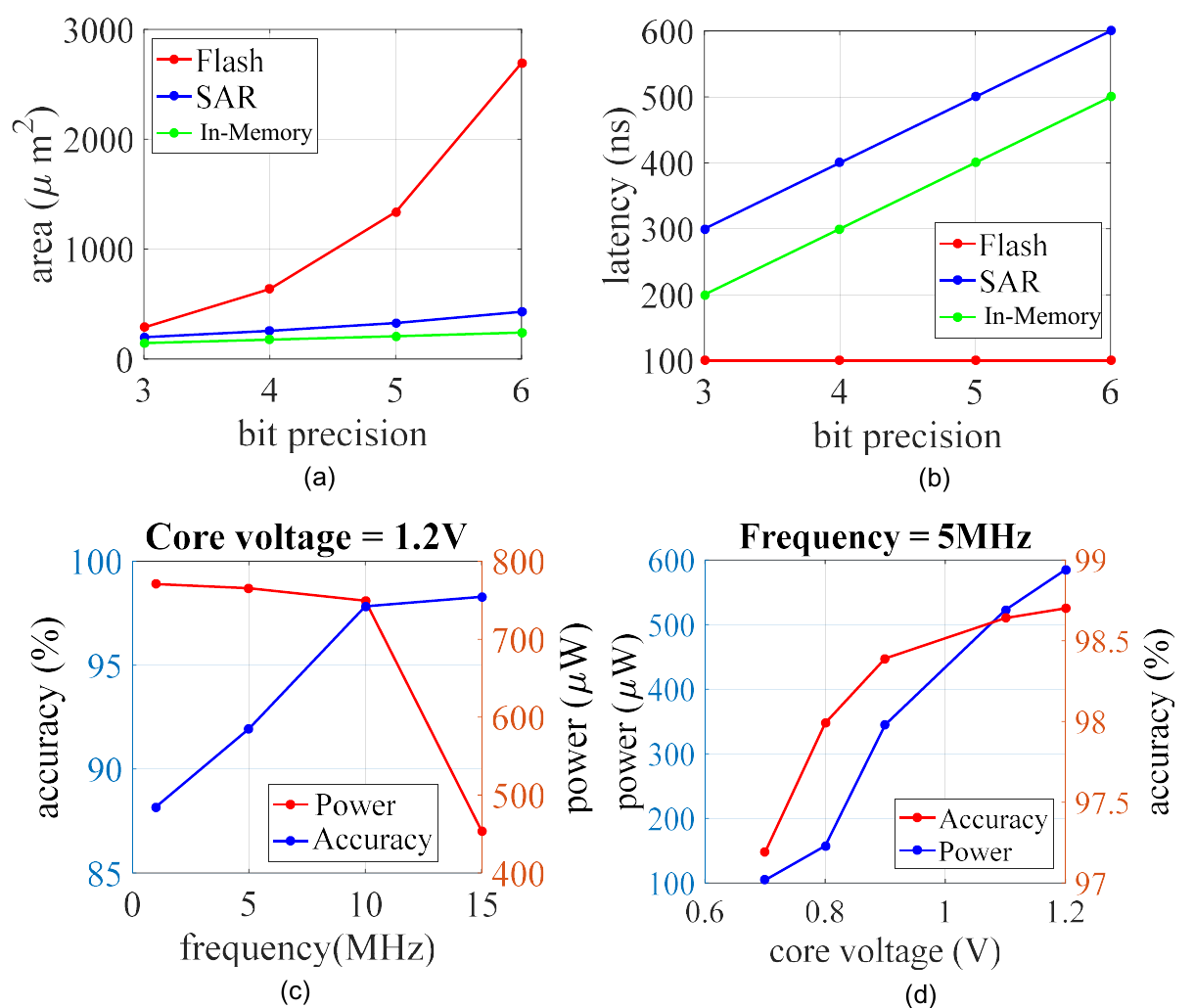}
  \caption{\textbf{Design space exploration of memory-immersed ADC:} Area in (a) and latency in (b) \textit{vs.} bit precision for different ADC styles. For memory-immersed ADC MNIST prediction accuracy and power \textit{vs.} frequency in (c) and \textit{vs.} operating voltage in (d). }
  \label{fig:design_space}
\end{figure}
\bibliographystyle{IEEEtran}
\bibliography{main.bib}
\end{document}